\documentclass[fleqn,10pt]{wlscirep}
\usepackage[utf8]{inputenc}
\usepackage[T1]{fontenc}
\usepackage{float} 

\usepackage{booktabs}       
\usepackage[table]{xcolor}  
\usepackage{multirow}       
\usepackage{array}          
\usepackage{graphicx}       
\usepackage{threeparttable}
\usepackage{nicematrix}

\definecolor{lightblue}{RGB}{220,230,241} 
\definecolor{headerblue}{RGB}{184,204,228} 
\definecolor{boxblue}{RGB}{163,186,221} 

\title{Deep Binarized Photonic Reservoir Computing for Ultrafast Multimedia Signal Processing}

\author[1,2,*]{Muhammad Waqar Iqbal}
\author[1,2]{Mohamad Alassir}
\author[1,2]{Nicolas Marsal}
\author[1,2,*]{Damien Rontani}
\affil[1]{Université de Lorraine, CentraleSupélec, LMOPS Laboratory, F-57070 Metz, France}
\affil[2]{Chaire Photonique, LMOPS Laboratory, CentraleSupélec, F-57070 Metz, France}
\affil[*]{muhammad-waqar.iqbal@centralesupélec.fr - damien.rontani@centralesupelec.fr}


\begin{abstract}
 We present a deep photonic neural network architecture based on ultrafast binary optical modulation from a digital micro-mirror device (DMD), optical scattering in random medium,  high-speed photodetection with a CMOS sensor, and time-multiplexed deep layer structure. Operating at Gigabit-per-second (Gb/s) processing rates, our system based on the reservoir computing (RC) framework achieves state-of-the-art performance across various multimedia tasks, including video, image and speech recognition. We show that the careful optimization of key physical intra- and inter-layer hyper-parameters can significantly enhance the deep photonic RC system ability to extract relevant temporal and spatial features via balancing memory retention and dynamical response of individual layers. This approach paves the way for highly scalable hierarchical photonic reservoir computing systems for high-throughput real-time multimedia signal processing.
\end{abstract}

\begin{document}

\flushbottom
\maketitle

\thispagestyle{empty}

\section{Introduction}
Reservoir computing (RC) is an innovative machine learning framework for efficiently training recurrent neural networks (RNNs). It uses a fixed and randomly initialized input layer, an untrained interconnected neural network ("the so-called reservoir") and a trainable output layer \cite{maass2002real,jaeger2004}. The reservoir acts as a high-dimensional nonlinear dynamical system that projects input data into feature representations more suitable for learning. In conventional reservoir computing, training is typically restricted to a linear readout layer that maps reservoir states to the desired output, significantly reducing computational cost \cite{lukosevicius2012}. More advanced implementations may employ nonlinear readout layers to further enhance expressive capacity \cite{antonik2017online}. Overall, this approach enables RC to achieve strong expressive power while maintaining a simplified learning process compared to traditional RNNs, with high performance across a wide range of machine learning tasks \cite{tanaka2019recent}.

Multiple physical platforms have been used to implement fast and energy-efficient RC systems, including analog/digital electronics \cite{appeltant2011information,du2017reservoir,sun2021sensor,haynes2015reservoir}, spintronics \cite{torrejon2017neuromorphic,jiang2019physical}, and photonics \cite{van2017advances,lugnan2020}. Among the most widely studied physical reservoir architectures, two main categories can be identified: (i) spatiotemporal reservoirs and (ii) single-node reservoirs with delayed feedback. Spatiotemporal reservoirs consist of discretely interconnected nonlinear devices forming a scalable network. Single-node reservoirs rely on a single nonlinear element with a delayed feedback loop to generate a time-multiplexed virtual high-dimensional space while reducing hardware complexity\cite{yan2024emerging}.

Photonics has enabled a wide range of high-performance physical implementations of reservoir computing (RC), spanning fiber-based oscillators \cite{duport2012all,larger2012PRL,larger2012Opex,larger2017high,Butschek2022}, semiconductor lasers with optical feedback \cite{brunner2013parallel,Vatin2019,bueno2021,Harkhoe2020}, photonic integrated circuits \cite{vandoorne2014experimental,takano2018compact,Sunada2019,ma2023integrated}, and free-space architectures based on spatial light modulation (SLM) \cite{rafayelyan2020large,bueno2018,antonik2019human,dong2019optical}. Among these approaches, SLM-based implementations stand out for their scalability and ability to exploit optical wavefront parallelism, enabling high-throughput and energy-efficient computation. However, their operation is typically limited to tens to a few hundred Hz due to the slow response time of liquid crystal on silicon (LCoS) technology, which constrains their applicability to high-speed processing tasks.

In parallel, alternative photonic hardware platforms have been explored to overcome the limitations of SLM-based systems. Digital micromirror devices (DMDs), for example, provide significantly higher modulation speeds by rapidly switching micromirror arrays between discrete ON/OFF states. While this binary operation constrains direct grayscale or high-precision modulation, recent studies have proposed binarized encoding approaches to better exploit DMD-based architectures for optical computing tasks \cite{dong2019optical}. Moreover, grayscale patterns can be approximated using pulse-width modulation or dithering techniques, and both phase and amplitude modulation can be achieved through the superpixel method, as demonstrated in \cite{Goorden2014Active}. Nevertheless, the effectiveness of these strategies for complex machine learning tasks involving high-dimensional multimedia signals remains largely unexplored, leaving open questions regarding their full computational potential.

Beyond hardware considerations, conventional single-layer reservoir computing architectures also exhibit intrinsic functional limitations, including trade-offs between non-linearity and memory capacity \cite{dambre2012information}, limited capacity to preserve long-range nonlinear temporal dependencies, and difficulties in processing hierarchical or multi-scale temporal dynamics \cite{moon2021hierarchical}. These constraints restrict their performance on more complex machine learning tasks. To address these limitations, hierarchical extensions known as deep reservoir computing (deep RC) have been proposed, where multiple reservoirs are stacked with unidirectional connectivity \cite{Gallicchio2017,moon2021hierarchical}. Such deep architectures have demonstrated improved nonlinear memory \cite{Gallicchio2018} and enhanced performance across machine learning benchmarks \cite{Li2024}, attributed to increased dynamical richness and representational capacity \cite{Gallicchio2023}. Recent photonic implementations of deep RC, involving the utilization of semiconductor lasers and optical fiber systems have shown that deep RC can enable fully analog multi-layer processing with improved performance in tasks such as speech recognition and signal processing \cite{shen2023deep}. These developments are strongly motivated by the broader success of deep learning in conventional machine learning, where convolutional neural networks (CNNs) have demonstrated that hierarchical multi-layer processing is essential for extracting complex features and learning abstract representations \cite{lecun2015deep}, thereby driving growing interest in photonic implementations of deep neural architectures that combine the computational efficiency of optics with the representational power of deep learning.

While these deep architectures motivate more advanced photonic implementations, the performance of physical reservoir computing systems remains strongly constrained by their underlying architectural choices. Time-delay reservoir computing (TDRC) systems, while experimentally convenient, rely on time multiplexing, which fundamentally restricts throughput to sequential processing of a single data stream and introduces latency that limits real-time operation. A better alternative is free-space reservoir computing (FSRC), which provides a fundamentally different approach by exploiting spatial degrees of freedom to achieve true parallelism. Using diffractive optical elements or spatial light modulators, FSRC enables simultaneous processing of information in multiple channels, significantly improving computational speed and scalability \cite{brunner2013parallel,yildirim2024nonlinear}, making it more suitable for computationally demanding machine learning tasks requiring both high throughput and increased network capacity.

In this work, we propose a deep photonic reservoir computing architecture integrating a digital micromirror device (DMD), optical scattering in random media, and high-speed CMOS photodetection to enable ultrafast (Gb/s) hierarchical processing of multimedia signals for scalable real-time intelligent information processing. To mitigate the encoding constraints imposed by binary DMD operation, we incorporate advanced binarized encoding strategies, such as basket encoding introduced in\cite{dong2019optical}, into the proposed deep RC architecture, enabling richer input representations beyond conventional binarized modulation. By stacking multiple reservoir layers hierarchically, our architecture enables progressive spatiotemporal feature extraction, improved robustness to noise, and enhanced nonlinear processing capabilities. Combined with the ultrafast modulation speed of DMDs, this deep reservoir architecture achieves efficient high-throughput computation for multimedia machine learning tasks, including image, video, and speech processing. Our results demonstrate that deep binarized photonic RC can achieve state-of-the-art performance among hardware reservoir implementations while operating at Gb/s rates, highlighting its potential as a scalable platform for real-time photonic machine learning.

\section{Deep Binarized Reservoir Computing Architecture}
\subsection{Experimental Implementation}
\begin{figure}[!h]
    \centering
    \includegraphics[width=1.00\textwidth]{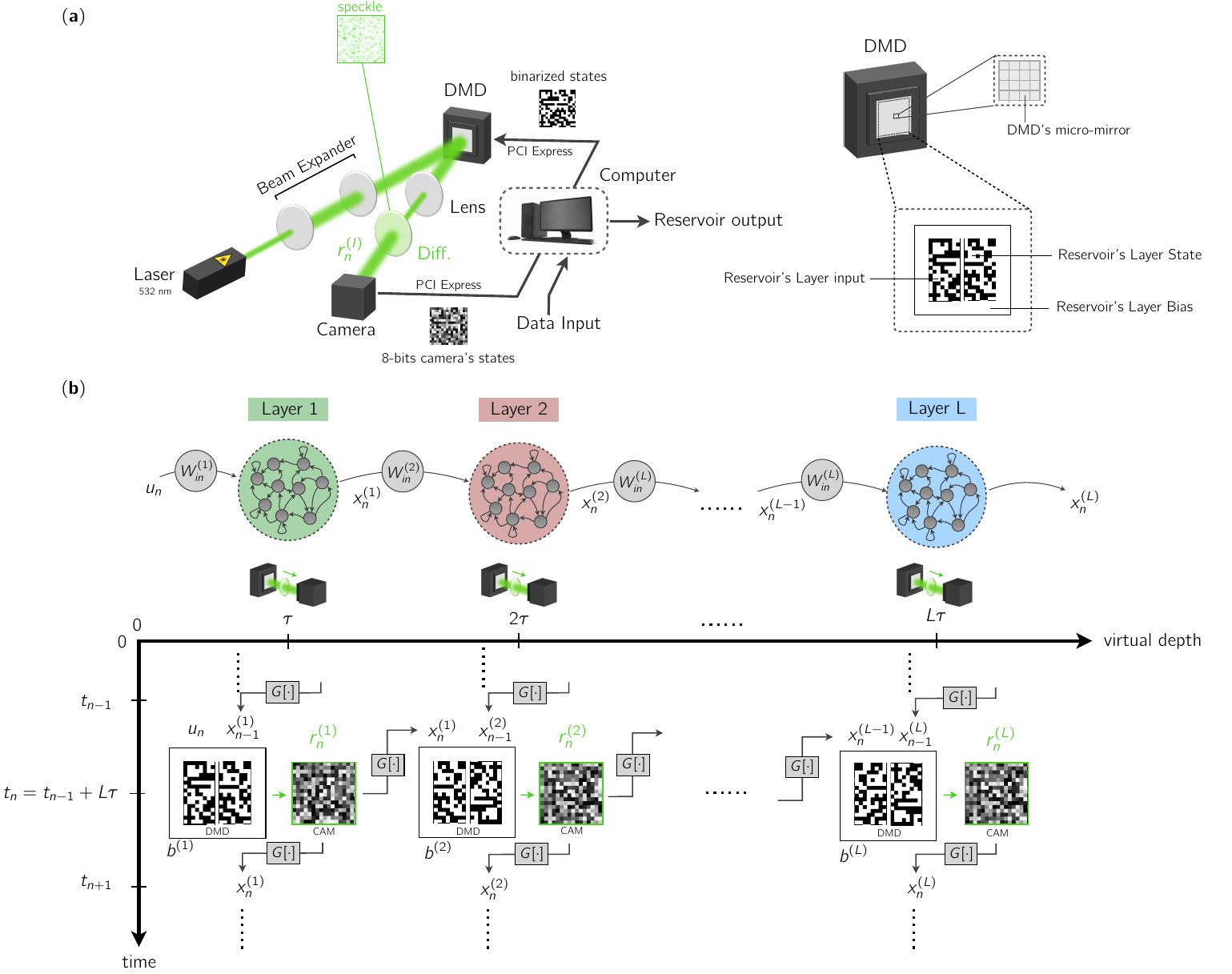}
    \caption{(a) Optical setup implementing the deep RC system. It consists of a 532-nm laser source, a digital micromirror device (DMD), and a ground-glass diffuser (Diff.) used to generate optical speckle patterns that are recorded by an $8$-bit monochrome high-speed camera. The discrete-temporal dynamics are produced by a digital feedback loop using PCI Express connectivity. The computer performs a basket encoding (8-bit to binarized 10 bits vector) to efficiently encode and map both the input data and the reservoir states onto the DMD. A conceptual representation of the locations of the reservoir input, states, and bias is also shown on the inset figure on the right. (b) Schematic illustration of the time virtualization strategy used in the deep RC architecture. The setup is used sequentially $L$ times to emulate the network's depth $L$. It leads  to a total duration of $L\tau$, where $\tau$ is the digital clock period of the system. This process yields one full time-step iteration of the deep RC. The state $x_n^{(l)}$ of the $l$-th layer $(l=1,\dots,L)$ at time $t_n$ is obtained via basket encoding of the optical speckle pattern corresponding to the reservoir state $r_{n}^{(l)}$, and is subsequently stored and propagated to the next deep RC iteration at $t_{n+1} = t_n + L\tau$.}
    \label{Fig_1_optical_setup_and_schematic_illustration}
\end{figure}

In this section, we describe the experimental implementation of the proposed deep reservoir computing (deep RC) architecture. Figure \ref{Fig_1_optical_setup_and_schematic_illustration} presents both the optical setup and a schematic illustration of the deep time-multiplexed RC architecture. The experimental system consists of a 532-nm laser source, a digital micromirror device (DMD) (ViALUX - V-9502c), a ground-glass diffuser (GGD), and a high-speed monochrome CMOS camera. The expanded and collimated laser beam uniformly illuminates the full active area of the DMD, where the input data and reservoir states are encoded as binary patterns. The reflected structured optical field is collected by a spherical lens and focused onto the GGD, which introduces multiple random scattering events and generates complex optical speckle patterns. These speckle patterns are recorded by the camera and provide the high-dimensional optical transformation underlying the reservoir dynamics, while intensity detection of the complex optical speckle field introduces the required nonlinearity through the modulus-square operation (see Fig. \ref{Fig_1_optical_setup_and_schematic_illustration}(a)).

To construct the reservoir state, only a subset of de-correlated macro-pixels, defined as spatially grouped neighboring camera pixels treated as individual state neurons, is selected from each captured speckle image and stored digitally. At the next iteration, the new input and the previous reservoir state are concatenated, binarized encoded, and projected onto the DMD, thereby creating the recurrent feedback required for reservoir dynamics. Since the DMD operates intrinsically in binary mode, an efficient binarized encoding strategy, namely basket encoding  \cite{dong2019optical}, is employed to map each 8-bit value into a 10-bit binary vector, enabling richer information representation while remaining compatible with the binary actuation (ON/OFF) of DMD micro-mirrors. A conceptual illustration of the encoded input, reservoir state, and bias regions is shown in the inset of Fig. \ref{Fig_1_optical_setup_and_schematic_illustration}(a), where a bias region refers to a dedicated area surrounding the encoded patterns that is kept in a fixed ON state, it contributes to the effective input–reservoir connectivity through optical scattering.

Figure \ref{Fig_1_optical_setup_and_schematic_illustration}(b) illustrates the implementation of the deep architecture. At each time step $t_n$, the encoded external input $u_n$, the previous encoded reservoir state $x_{n-1}^{(1)}$, and the layer-dependent bias term $b^{(1)}$ are simultaneously displayed on the DMD to compute the updated encoded state $x_n^{(1)}$ for the first layer. After optical propagation and camera acquisition, the resulting speckle pattern is used to extract the updated new reservoir state (i.e. $r_n^{(1)}$). This state is stored as an 8-bit representation for reservoir dynamics and, in parallel, converted into a binarized representation via basket encoding to be used in the subsequent input–state encoding cycle.

One full reservoir update requires a clock period $\tau$, determined by synchronization between the DMD, camera acquisition, and digital processing. To realize a deep RC with $L$ layers, the same physical system is sequentially reused $L$ times in a time-multiplexed manner. Specifically, the binarized output state of layer $l-1$, denoted $x_n^{(l-1)}$, is fed as the input to layer $l$, yielding the updated binarized encoded state $x_n^{(l)}$. Consequently, one full deep reservoir update requires a total duration of $L\tau$. This temporal virtualization strategy enables the implementation of hierarchical reservoir depth without requiring physical duplication of optical components such as the DMD, diffuser, and camera.

Only the first reservoir layer directly receives the external input, while subsequent layers process progressively transformed reservoir states. As a result, the influence of the original input is gradually attenuated with depth, enabling the emergence of hierarchical internal representations. To control information propagation across layers, layer-dependent leakage rates $\alpha^{(l)}$ are introduced within the state update dynamics. 
The leakage parameter governs the trade-off between temporal memory retention and dynamical responsiveness, where lower values emphasize slower state evolution and higher values increase sensitivity to recent temporal variations.

In addition, the bias term $b^{(l)}$ is independently adjusted for each layer by modulating the proportion of fixed ON pixels on the DMD. This helps prevent over-dominance of bias-driven dynamics, stabilizes the reservoir evolution, and maintains sufficient diversity in the generated speckle patterns across layers. Since the bias modifies the illumination pattern incident on the scattering medium, it effectively leads to different input–reservoir mappings for each layer, despite using the same physical optical system. 

\subsection{Theoretical Model}
Here, we present a detailed theoretical modeling framework for our proposed deep binarized photonic RC. As described above, the architecture consists of multiple reservoir layers stacked hierarchically: Only the first layer receives the external input, while each subsequent layer processes the output state of the previous one. The state transition function governing the deep RC dynamics is then formulated as follows
\begin{equation}
    r^{(l)}_{n} = (1 - \alpha^{(l)}) r^{(l)}_{n-1} + \alpha^{(l)} f\left(W_{in}^{(l)} G[u^{(l)}_{n}] + W_{res}^{(l)} G[r^{(l)}_{n-1}] + W_{b}b^{(l)} \right),
\end{equation}
where \(G[r_n^{(l)}] =  x^{(l)}_n \) represents the binarized reservoir state at time step \( t_n \) for the \( l \)-th layer obtained from the basked encoding function $G(\cdot)$ applied to the optical intensity speckle (stored as a reservoir state i.e. $r_n^{(l)}$) detected during the $l$-th use of hardware resource. $u^{(l)}_n$ is the $l$-th reservoir layer input and reads:
\begin{equation}
    u^{(l)}_n = 
        i_n \text{ if } l = 1 \quad \text{ and } \quad u_n^{(l)} = x^{(l-1)}_n \text{ if } l > 1
\end{equation}
with $i_n$ representing the external input data at $n$-th iteration. The parameter $\alpha^{(l)}$ denotes the leakage rate associated with the $l$-th reservoir layer in the deep RC architecture. It governs the balance between memory retention and state update dynamics. In deep reservoir architectures, this parameter can be kept constant or varied across layers to control temporal processing. In this work, we adopt a depth-dependent scheduling strategy to better separate short- and long-term temporal representations. Specifically, layer-dependent leakage rates are employed to induce diverse temporal dynamics across the network, enabling different layers to capture information over multiple timescales and balance memory retention with responsiveness to incoming inputs and nonlinear transformations. The layer-wise leakage rate is defined as a linear function of depth:
\begin{equation}
\alpha^{(l)} = \alpha_{\min} + \frac{(\alpha_{\max} - \alpha_{\min})(l-1)}{L-1}, \quad l \in [1,L],
\label{eq:leakage_rate_variation}
\end{equation}
where $L$ is the total number of reservoir layers in the deep RC architecture, and $\alpha_{\min}$ and $\alpha_{\max}$ define the leakage range across the network depth. 
The matrices \( W_{in}^{(l)} \) and \( W_{res}^{(l)} \) refer to the input connectivity and the reservoir connectivity matrices for the $l$-th layer, respectively. In our experimental setup, these matrices are physically implemented using a ground glass diffuser in combination with the bias term $b^{(l)}$, which makes them static (and different) for the $L$ layers within the deep RC. The function $f$ denotes the nonlinear activation function and in our Deep photonic RC architecture, the light intensity measured by the camera constitutes the only source of nonlinearity. The final output of the deep photonic RC system is computed as
\begin{equation}
    Y_{\text{RC},n} = W_{\text{out}} \big[ r^{(1)}_n, r^{(2)}_n, \dots, r^{(L)}_n \big]^T.
\end{equation}
Here, $r^{(l)}_n\in \mathbb{R}^{N_{l}}$  denotes the reservoir state vector at time step $t_n$ from the $l$-th photonic reservoir layer with $N_l$ the associated number of neurons. The concatenated vector  $R_n \in \mathbb{R}^{N_X}$ represents the entire system's state across all layers with $N_X = \sum_{l=1}^L N_l$. The readout weight matrix $W_{\text{out}}\in  \mathbb{R}^{N_Y \times N_X}$  maps the transposed reservoir state vector to the final output $Y_n \in \mathbb{R}^{N_Y}$, where $N_Y$ is the number of output classes. The training of the RC system is carried out in a supervised learning framework, where the goal is to optimize  $W_{\text{out}}$  such that the system's output approximates the desired target responses. During training, $T$ input samples are presented to the reservoir, and the corresponding reservoir states are collected into the matrix $\mathbf{R}_{0:T-1} \in \mathbb{R}^{N_X \times T} = [R_0,\dots,R_{T-1}]$. The associated target outputs form the matrix $\mathbf{Y}_{0:T-1} \in \mathbb{R}^{N_Y \times T}$ (one hot encoded labels for each output class). The optimal readout weights are obtained by solving the ridge regression problem:

\begin{equation}
    W_{\text{out}} = \arg\min_W \, \| W \mathbf{R}_{0:T-1} - \mathbf{Y}_{0:T-1} \|_2^2 + \lambda \| W \|_2^2,
\end{equation}
where \( \lambda \) is the regularization parameter, whose value is determined through Bayesian optimization to ensure robust generalization and optimal performance.

\section{Experimental Results}
\subsection{Benchmark Performance}
\begin{figure}[h!]
    \centering
    \includegraphics[width=1.0\linewidth]{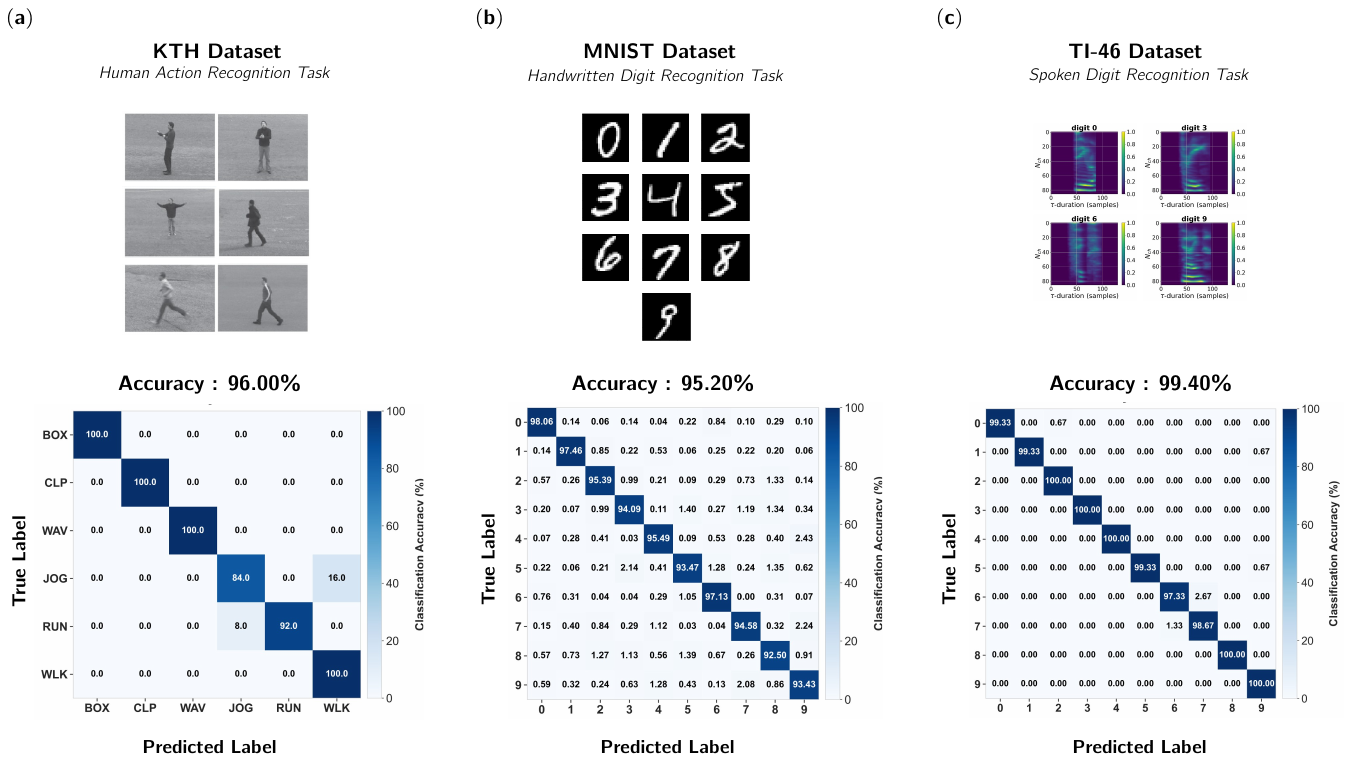}
    \caption{Deep photonic reservoir computing (RC) performance across three multimedia signal processing tasks: (a) human action recognition using the KTH dataset, (b) handwritten digit classification using the MNIST dataset, and (c) spoken digit classification using the TI-46 dataset. Confusion matrices report mean classification accuracies over three independent experiments for five-layer deep RC architectures employing 10000 neurons for KTH, 3,500 neurons for MNIST, and 500 neurons for TI-46. The results demonstrate that the proposed deep photonic RC achieves classification accuracies exceeding 95\% across all tasks. Representative input samples are displayed above each confusion matrix, including video frames from KTH, handwritten digit images from MNIST, and time-frequency cochleagrams derived from TI-46 audio recordings.}
    \label{fig:Deep_RC_Summarized_results}
\end{figure}

In this section, we evaluate the performance of the proposed deep photonic RC architecture on classification tasks spanning multiple multimedia signal modalities. Specifically, we assess the system on three representative machine learning benchmarks: (i) human action recognition using the KTH video dataset, (ii) handwritten digit classification using the MNIST image dataset, and (iii) spoken digit recognition using the TI-46 audio dataset. A summary of the obtained results is presented in Fig. \ref{fig:Deep_RC_Summarized_results}, while a comparative analysis with existing digital and hardware-based implementations is provided in Table \ref{table:performance_comparison}.

\begin{table}[h!]
\centering
\small
\renewcommand{\arraystretch}{0.5} 
\caption{Performance comparison of the proposed deep binarized photonic reservoir computing system with representative state-of-the-art digital and hardware-based approaches on KTH, MNIST, and TI-46 benchmarks.}
\label{table:performance_comparison}
\begin{threeparttable}
\begin{NiceTabular}{|>{\centering\arraybackslash}m{1.1cm}|>{\centering\arraybackslash}m{3.2cm} >{\centering\arraybackslash}m{5.0cm} >{\centering\arraybackslash}m{2cm} >{\centering\arraybackslash}m{2cm} >{\raggedright\arraybackslash}m{2cm}|}
\hline
\textbf{Task} & \textbf{Reference} & \textbf{Method \& Preprocessing} & \textbf{Network Size} & \textbf{Speed (fps)} & \textbf{Accuracy (\%)} \\
\hline

\Block{7-1}{\centering KTH} 
& Jaouedi et al. (2020)\cite{jaouedi_new_2020} & GRNN + motion tracking + GMM/KF & – & – & \shortstack[l]{FD 96.30 \\ S1 --} \\
& Sharif et al. (2017)\cite{sharif_framework_2017} & Multi-class SVM + HOG + LBP + texture features & – & 10 & \shortstack[l]{FD 99.30 \\ S1 --} \\
& Khan et al. (2021)\cite{khan_resource_2021} & 26-layer CNN + PDaUM preprocessing & $\sim$195M & – & \shortstack[l]{FD 98.30 \\ S1 --} \\
& Shu et al. (2014)\cite{shu_bio-inspired_2014} & Bio-inspired SNN + preprocessing & 24,000 & – & \shortstack[l]{FD 92.30 \\ S1 95.30} \\
& Picco et al. (2023)\cite{picco_high_2023} & TDRC + silhouette segmentation, keyframe selection, HOG, PCA & 600 & 160 & \shortstack[l]{FD 90.83 \\ S1 96.67} \\
& Antonik et al. (2019)\cite{antonik2019human} & Photonic RC + HOG/PCA & 16,384 & 2–7 & \shortstack[l]{FD -- \\ S1 91.30} \\
& \textbf{This work} & \textbf{Deep binarized photonic RC + HOG/PCA} & \textbf{10,000} & \textbf{$\sim$1000} & \shortstack[l]{\textbf{FD --} \\ \textbf{S1 96.00}} \\
\hline

\Block{6-1}{\centering MNIST} 
& Yepes et al. (2017)\cite{yepes_improving_2017} & SNN deployed on IBM TrueNorth chip & – & – & 98.00 \\
& Wang et al. (2025)\cite{wang_asymmetrical_2025} & Deep photonic NN + PCA reduces 784 features to 8 & – & – & 95.80 \\
& Du et al. (2017)\cite{du2017reservoir} & Memristor-based RC hardware & 88 & – & 88.10 \\
& Talukder et al. (2025)\cite{talukder_spiking_2025} & Spiking photonic NN trained on $\sim$6000 images & 40,000 & – & 83.50 \\
& Han et al. (2020)\cite{han_hardware_2020} & SNN implemented on FPGA & 16,384 & 161 & 97.06 \\
& \textbf{This work} & \textbf{Deep binarized photonic RC + HOG/PCA} & \textbf{3500} & \textbf{$\sim$1000} & \textbf{95.20} \\
\hline

\Block{6-1}{\centering TI-46} 
& Zheng et al. (2021)\cite{zheng_parameters_2021} & TDRC + nonlinear Duffing oscillator + Lyon cochleagram & 100 & – & 99.80 \\
& Picco et al. (2025)\cite{picco_deep_2025} & Deep photonic RC + CMA-ES evolutionary algorithm + Lyon cochleagram & 600 & – & 99.40 \\
& Paquot et al. (2012)\cite{paquot2012optoelectronic} & Optoelectronic RC + Lyon cochleagram & 200 & – & 99.60 \\
& Wan et al. (2025)\cite{wan_reservoir_2025} & TDRC system based on polymer electrolyte-gated MoS$_2$ transistors & – & – & 95.10 \\
& Zhong et al. (2021)\cite{zhong2021dynamic} & Dynamic memristor-based RC + Lyon cochleagram & 400 & – & 99.60 \\
& \textbf{This work} & \textbf{Deep binarized photonic RC + Lyon cochleagram} & \textbf{500} & \textbf{$\sim$1000} & \textbf{99.40} \\
\hline
\end{NiceTabular}

\begin{tablenotes}
\footnotesize
\item Network size and processing speed are reported as provided in the cited works. Missing values were not reported.
\item Abbreviations — FD: Full Dataset; S1: Scenario 1 (KTH); HOG: Histogram of Oriented Gradients; PCA: Principal Component Analysis; CNN: Convolutional Neural Network; SNN: Spiking Neural Network; RC: Reservoir Computing; PIC: Photonic Integrated Circuit; FPGA: Field-Programmable Gate Array; CMA-ES: Covariance Matrix Adaptation Evolution Strategy.
\end{tablenotes}
\end{threeparttable}
\end{table}

Across all benchmark tasks, the proposed deep photonic RC architecture achieves highly competitive performance, reaching approximately $96 \%$ on KTH, $95.2 \%$ on MNIST, and $99.4 \%$ on TI-46 (see Fig. \ref{fig:Deep_RC_Summarized_results}), while operating at ultrafast processing speeds of approximately 1000 fps per layer. 

As summarized in Table~\ref{table:performance_comparison}, the proposed system achieves competitive performance compared to existing neuromorphic hardware and photonic reservoir computing implementations, while maintaining high classification accuracy, favorable processing throughput, and reduced network complexity with compact reservoir sizes. While several prior approaches report slightly higher accuracies, these methods often rely on substantially larger trainable models or computationally intensive preprocessing pipelines. For example, the KTH task performance reported in Ref.~\cite{picco_high_2023} relies on multiple preprocessing and feature-engineering stages, including keyframe subsampling, silhouette extraction, and feature reduction prior to reservoir inference, enabling operation with a smaller reservoir size. In contrast, our approach achieves comparable performance while maintaining ultrafast processing speeds and moderate network complexity within a streamlined HOG/PCA-based pipeline. Similarly, on the MNIST task, competing approaches frequently depend on large-scale deep or spiking neural architectures~\cite{yepes_improving_2017,han_hardware_2020,wang_asymmetrical_2025}, while our system attains competitive accuracy using a compact deep photonic reservoir operating in real time. For the TI-46 task, the proposed architecture achieves performance comparable to state-of-the-art hardware RC systems using only standard Lyon cochleagram preprocessing and a moderate reservoir size. Furthermore, direct efficiency comparisons remain challenging for several previously reported systems due to the absence of reported throughput or network-size metrics. Overall, these results highlight the potential of deep photonic RC as a scalable, hardware-efficient, and high-throughput platform for real-time multimedia neuromorphic computing.

The dataset-specific preprocessing pipelines are detailed in the Methods section. To optimize system performance, we performed a systematic exploration of the deep RC design space, including the number of reservoir layers ($L=2$ to $5$), neuron distributions across layers, layer-wise leakage rate scheduling, and inter-layer bias optimization. These architectural and hyperparameter studies were found to be essential for achieving robust performance across heterogeneous tasks. The following section presents a detailed parametric analysis highlighting the role of these key design parameters and providing practical guidelines for tuning deep RC architectures across different machine learning tasks.

\subsection{Parametric Sensitivity Analysis}
In this section, we perform a detailed parametric sensitivity analysis on our system using the TI-46 spoken digit recognition benchmark. We chose this dataset because it exhibits rich multi-timescale temporal dynamics, containing both short-term acoustic features (e.g., phonetic transitions and spectral variations) and long temporal dependencies associated with the sequential structure of spoken digits. Such characteristics make TI-46 a particularly relevant benchmark for evaluating the ability of deep binarized photonic RC architecture to hierarchically process and retain temporal information across multiple layers. Details of the TI-46 dataset, as well as the preprocessing pipeline and training procedure, are provided in the Methods section.

\subsubsection*{Effect of Neuron Distribution Across Layers} 
We first investigate the impact of neurons allocation across layers in the deep RC architecture. Three layer-wise neurons allocation strategies are considered: (i) decreasing neurons allocation with increasing depth, (ii) uniform allocation of neurons across all layers, and (iii) increasing neuron allocation with increasing depth, corresponding to the reverse ordering in (i). The decreasing neurons allocation strategy follows a normalized power-law distribution under a fixed total neuron budget, with implementation details provided in the Methods section.

\begin{figure}[h!]
    \centering
    \includegraphics[width=0.98\linewidth]{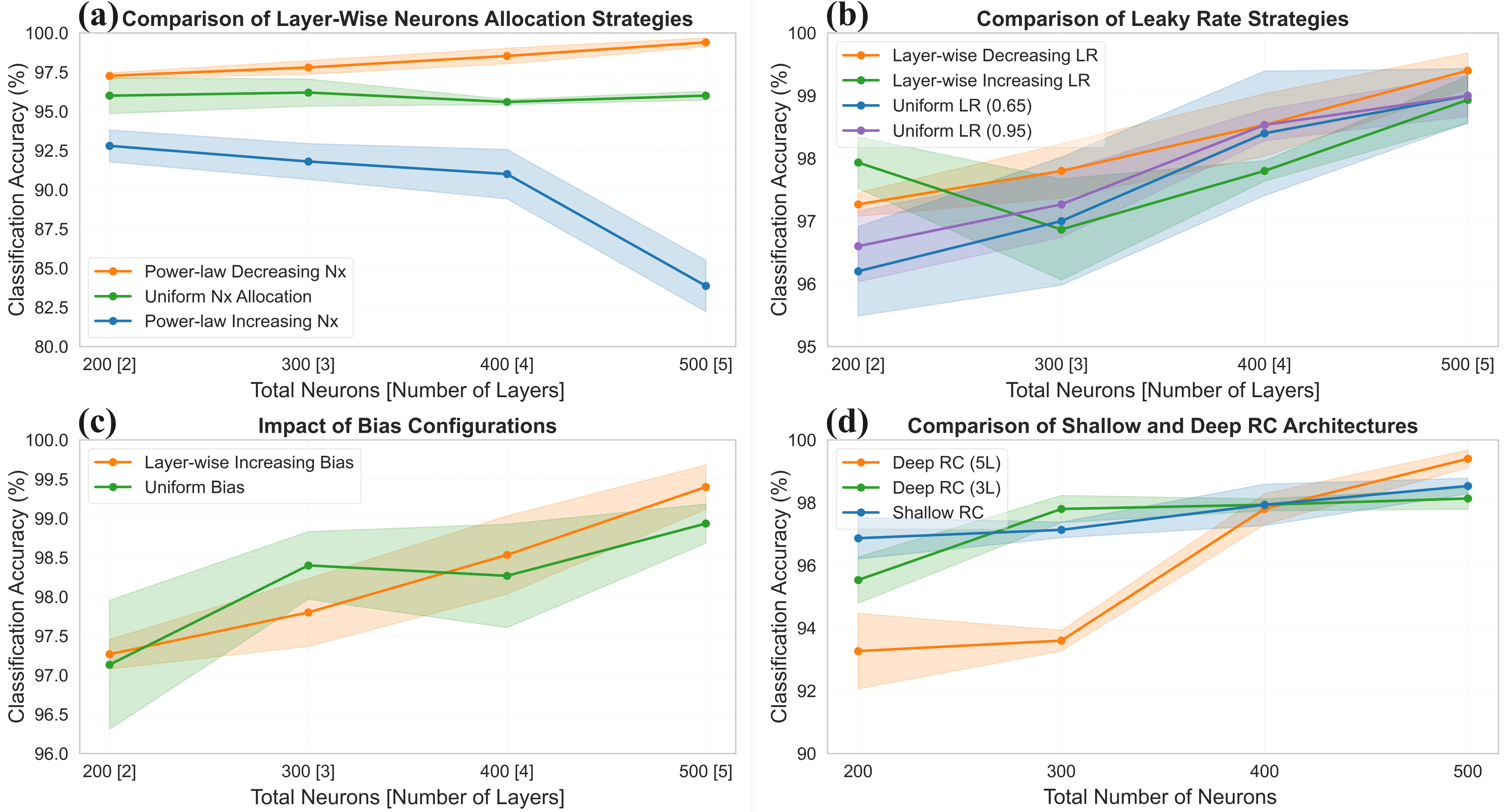}
    \caption{Ablation studies on key architectural hyperparameters in deep reservoir computing (RC) for the TI-46 audio recognition task, evaluating their impact on classification accuracy across network depths of 2 to 5 layers while maintaining a fixed total neuron budget of $N = 100 \times L$. Shaded regions represent confidence intervals based on standard deviation over multiple experimental runs (three experiments per configuration, each using 10-fold cross-validation). (a) Comparison of neurons distribution strategies under a linearly decreasing leakage rate configuration (from $\alpha_{max} = 0.95$ to $\alpha_{min} = 0.65$ via Eq. \ref{eq:leakage_rate_variation}) and a mild layer-wise increasing bias profile; decreasing neurons allocation via power-law decay (orange), uniform allocation of 100 neurons per layer (green), and increasing neurons allocation obtained by reversing the power-law decay profile (blue). (b) Effect of leakage rate configurations under decreasing neurons allocation strategy and mild layer-wise increasing bias profile across network depths of 2 to 5: linearly decreasing leakage rates from $\alpha_{max} = 0.95$ to $\alpha_{min} = 0.65$ (orange), linearly increasing leakage rates from $\alpha_{max} = 0.65$ to $\alpha_{min} = 0.95$ (green), and fixed leakage rates of $\alpha = 0.65$ (blue) and $\alpha = 0.95$ (purple) across layers. (c) Influence of bias configurations under decreasing neurons allocation strategy and linearly decreasing leakage rates from $\alpha_{max} = 0.95$ to $\alpha_{min} = 0.65$ via Eq. \ref{eq:leakage_rate_variation}: layer-wise increasing bias profile in the shallow layers (orange) versus uniform bias across all layers in the deep RC architecture (green). (d) Performance of shallow RC (i.e., a single-layer RC (blue) with all neurons in the input layer and fixed $\alpha = 0.95$) compared to deep RC with network depths of three (green) and five layers (orange), under decreasing neurons allocation strategy and mild layer-wise increasing bias profile, highlighting the benefits of hierarchical temporal representations in deeper architectures.}
    \label{fig:Parametric_Analysis_KTH_task}
\end{figure}

Figure~\ref{fig:Parametric_Analysis_KTH_task}(a) shows the classification performance of deep RC architecture with depths ranging from 2 to 5 layers under the three neuron allocation strategies. In all experiments, the total number of neurons is kept constant, while a layer-wise decreasing leakage schedule is applied according to Eq.~\ref{eq:leakage_rate_variation}. A mild layer-wise increasing bias profile is also introduced by progressively increasing the fraction of DMD micromirrors maintained in a fixed ON state across deeper layers.

The results clearly show that allocating more neurons to earlier layers and progressively fewer neurons to deeper layers yields the highest classification accuracy across all network depths. This behavior arises from the particular structure of the proposed deep RC where only the first reservoir layer directly receives the external input and therefore require a bigger representational capacity to encode data features. In contrast, deeper layers process progressively transformed internal reservoir states, thus assigning larger neuron budgets in later layers is less efficient computationally and more prone to propagating redundant or noisy representations. This qualitative interpretation is consistent with the lowest performance observed with the increasing allocation strategy, where deeper layers contain more neurons (see blue curve in  Fig.~\ref{fig:Parametric_Analysis_KTH_task}(a)).

The decreasing neurons allocation strategy also aligns with the hierarchical processing mechanism of deep RC, where early layers capture rapidly varying input features and deeper layers progressively refine and integrate higher-level temporal representations. Such a hierarchy is particularly advantageous for the TI-46 spoken digit recognition task, where each utterance contains both rapidly varying local acoustic transitions and slower temporal dependencies distributed across the full digit utterance. Overall, these results indicate that decreasing neuron allocation strategy improves both computational efficiency and hierarchical spatiotemporal processing in our deep binarized photonic RC system.

\subsubsection*{Effect of Leakage Rate Variation}
The leakage parameter $\alpha$ controls the temporal integration dynamics of reservoir neurons and therefore directly influences the memory properties of the deep RC architecture. To evaluate its effect, we compare four layer-wise leakage rate configurations: (i) a fixed leakage rate across all layers with $\alpha = 0.65$, (ii) a fixed leakage rate with $\alpha = 0.95$, (iii) a linearly decreasing leakage rate schedule across depth, defined by Eq.~\ref{eq:leakage_rate_variation} with $\alpha_{\max}=0.95$ and $\alpha_{\min}=0.65$, and (iv) a linearly increasing leakage rate schedule using the reverse ordering in (iii). All experiments are conducted using the decreasing neuron allocation strategy described in the previous subsection, while keeping the same layer-wise increasing bias profile across depth.

Figure~\ref{fig:Parametric_Analysis_KTH_task}(b) shows that the impact of leakage rate scheduling is more pronounced in deep RC architectures with up to three layers, where clear performance differences are observed between the tested configurations. In this regime, the choice of leakage profile has a stronger influence on the reservoir dynamics, leading to noticeable variations in classification accuracy. However, as the network depth increases beyond three layers, the performance gap between different leakage strategies progressively reduces. For four- and five-layer architectures, all configurations converge toward similar performance levels, suggesting a partial saturation of the benefit provided by leakage tuning in deeper hierarchies.

For the TI-46 task, the linearly decreasing leakage rate schedule consistently achieves the most favorable performance, particularly in deeper architectures. In this configuration, higher leakage values in the first layer enable rapid adaptation to incoming acoustic features, which is important for capturing fast local variations in speech signals. In contrast, lower leakage values in deeper layers promote slower temporal integration, allowing these layers to accumulate and refine higher-level temporal structure over longer time horizons.

Importantly, the effectiveness of a given leakage schedule is task dependent. Although not explicitly shown here, additional experiments on datasets such as KTH and MNIST indicate that alternative leakage configurations can yield comparable or slightly improved performance, reflecting their different temporal and spatial complexity profiles. Overall, these results indicate that no single leakage scheduling strategy is universally optimal. Instead, leakage rate should be treated as a task-dependent hyperparameter that controls the balance between short-term responsiveness and long-term temporal integration across the reservoir hierarchy. 

\subsubsection*{Effect of Bias Configuration Across Layers}
In this parametric study, we further investigate the influence of layer-wise bias configuration on the performance of the proposed deep photonic RC architecture. As introduced in the experimental implementation, the bias is defined as the fraction of DMD mirrors maintained in the ON state throughout each input presentation. This parameter controls the baseline optical intensity level and influences the operating regime of the reservoir by modifying the speckle field statistics and, consequently, the resulting reservoir state dynamics.

Two bias configurations are considered: (i) a mild layer-wise increasing bias profile, in which the fraction of fixed ON micromirrors is gradually increased in deeper layers, and (ii) a uniform bias applied identically across all layers. The motivation for introducing an increasing bias profile is to compensate for the progressively reduced direct influence of the external input in deeper reservoir layers, thereby maintaining sufficient activation levels and preserving the effective dynamic range across the hierarchical architecture.

All experiments are performed on deep RC architectures with depths ranging from 2 to 5 layers. Neurons are distributed according to the power-law allocation strategy described in the Methods section, while the leakage rate follows a linearly decreasing schedule from $\alpha_{max} = 0.95$ to $\alpha_{min} = 0.65$ via  Eq.~\ref{eq:leakage_rate_variation}. These configurations correspond to the favorable operating conditions identified in the previous parametric analyses.

Figure~\ref{fig:Parametric_Analysis_KTH_task}(c) shows that for shallow and intermediate architectures (up to four layers), both bias configurations yield comparable classification performance, indicating that the system remains relatively insensitive to bias modulation at moderate depths. However, for deeper architectures with five layers, the mild increasing bias configuration consistently provides a small but systematic improvement in classification accuracy.

Interestingly, under the increasing bias strategy, performance continues to improve progressively with increasing depth, whereas the uniform bias configuration exhibits a comparatively earlier performance saturation, with only marginal gains beyond shallow architectures. This behavior suggests that layer-wise bias adaptation becomes increasingly relevant as the reservoir hierarchy deepens.

The observed improvement can be attributed to enhanced dynamical stability in deeper reservoir layers. Specifically, introducing a slightly elevated baseline optical activation helps prevent weak signal attenuation in later stages and preserves a richer distribution of speckle-induced reservoir states. As a result, information propagation across layers remains more stable, reducing progressive signal degradation and improving robustness of the hierarchical state representation.

Overall, these results indicate that bias plays a secondary but non-negligible role in our deep photonic RC system. Although its influence is less pronounced than neurons allocation, an appropriately designed layer-wise bias profile can improve dynamical stability and yield measurable performance gains in deeper reservoir architectures.

\subsubsection*{Effect of Network Depth (Shallow vs. Deep RC)}
To assess the influence of architectural depth on system performance, we compare a conventional shallow RC architecture consisting of a single recurrent layer with deep RC architectures comprising 3 and 5 reservoir layers. For a fair comparison, all configurations are evaluated under identical total neuron budgets.

In the shallow RC configuration, all neurons are concentrated in a single reservoir layer directly receiving the external input. In contrast, for our deep RC architectures, the total neuron budget is distributed across layers using the decreasing power-law allocation strategy introduced previously, where earlier layers receive a larger fraction of neurons. This allocation was shown in Fig.~\ref{fig:Parametric_Analysis_KTH_task}(a) to be the most favorable configuration for the proposed hierarchical architecture. In addition, deep RC architectures employ the optimized layer-wise decreasing leakage schedule defined in Eq.~\ref{eq:leakage_rate_variation}, together with a mild increasing bias profile across depth, as identified in the preceding analyses.

Figure~\ref{fig:Parametric_Analysis_KTH_task}(d) presents the classification accuracy as a function of the total neuron budget for shallow and deep configurations. For small computational budgets (approximately 200 total neurons), the shallow RC exhibits competitive performance. This behavior is expected, as all available neurons are directly allocated to the input-facing reservoir layer, maximizing immediate encoding capacity of the incoming signal. Under such constrained resource conditions, distributing neurons across multiple layers reduces the number of neurons available in the first layer, which can limit initial feature extraction.

However, as the total neuron budget increases, the advantages of hierarchical processing become progressively more pronounced. In particular, deep RC with 3 layers already surpasses the shallow RC for intermediate budgets (300--400 neurons), while the 5-layer deep RC achieves the highest overall performance once the total neuron budget reaches approximately 500 neurons. This transition indicates that deeper architectures begin to benefit from sufficient representational capacity in both early and deeper layers, allowing the system to exploit hierarchical temporal processing more effectively.

The improved performance of deeper architectures can be attributed to progressive feature transformation across layers. While early layers primarily encode rapidly varying input features, deeper layers operate on increasingly abstract internal reservoir representations, enabling richer temporal integration and improved modeling of long-range dependencies. Combined with optimized neuron allocation, leakage scheduling, and bias tuning, this hierarchical organization allows the deep RC to extract more informative spatiotemporal features than a shallow architecture under sufficiently large computational budgets.

Overall, this parametric sensitivity analysis demonstrate that architectural depth alone does not guarantee improved performance. Instead, the benefits of deep RC emerge only when the number of layers is carefully matched to the available neuron budget and accompanied by appropriate layer-wise hyperparameter optimization. For limited computational resources, shallow RC may remain preferable, however, under moderate-to-large neuron budgets, properly optimized deep RC architectures consistently outperform their shallow counterparts, highlighting the importance of jointly designing depth and resource allocation in our deep binarized photonic RC.

\section{Conclusion}  
In this work, we presented a deep binarized photonic reservoir computing architecture combining DMD-based binary optical modulation, optical random scattering, and high-speed CMOS detection within a time-multiplexed hierarchical reservoir framework. By integrating advanced binarized encoding strategies with multi-layer reservoir dynamics, the proposed system overcomes key limitations of conventional DMD-based photonic RC architectures for multimedia signal processing, particularly the restricted encoding resolution imposed by binary modulation. The proposed time-multiplexed deep architecture enables hierarchical processing of multimedia signals across successive reservoir layers while preserving high-throughput operation, achieving an effective information processing speed in the Gigabit-per-second (Gb/s).

The proposed architecture achieves state-of-the-art performance among hardware-based reservoir computing systems across multiple multimedia classification benchmarks, reaching classification accuracies of $99.40\%$ on the TI-46 spoken digit dataset, $95.20\%$ on MNIST handwritten digit recognition, and $96.00\%$ on KTH human action recognition. These results demonstrate strong generalization across both spatial and temporal data modalities while operating within a real-time photonic computing framework. Through systematic parametric analysis, we further showed that architectural depth, layer-wise neuron allocation, leakage scheduling, and bias optimization all play a critical role in shaping the dynamical properties and overall performance of deep binarized photonic RC system.

Beyond benchmark performance, this work highlights the potential of hierarchical photonic reservoir architectures for scalable and high-throughput intelligent signal processing. The combination of ultrafast DMD modulation, free-space optical parallelism, and deep reservoir dynamics provides an efficient hardware platform capable of robust feature extraction and temporal information processing with reduced hardware complexity. In particular, the proposed architecture demonstrates that deep binarized photonic RC can effectively bridge the gap between the computational efficiency of optical hardware and the representational advantages of deep learning-inspired architectures. Future work will investigate deeper reservoir hierarchies, hybrid photonic-electronic implementations to further improve scalability and applicability to real-world signal processing tasks. Overall, this study establishes deep binarized photonic reservoir computing as a promising direction toward next-generation high-speed neuromorphic computing systems.

\section*{Methods}
\subsection*{Basket Encoding}\label{sec:methods_basket_encoding}
To enable binary-compatible encoding on the DMD while preserving similarity relationships between nearby input values, we employ basket encoding, originally proposed by Dong \textit{et al.} for optical reservoir computing using multiple light scattering~\cite{dong2019optical}. This encoding maps scalar values into overlapping binary activation patterns, such that similar input values produce similar binary representations with small Hamming distances, while more distant values become progressively decorrelated.

Basket encoding is employed to map continuous-valued input and reservoir states onto the DMD, which intrinsically supports only binary modulation. It is defined as a transformation operator \(G(\cdot)\) that converts each scalar value of an input or reservoir state vector into a distributed binary representation. The action of basket encoding on each scalar value \(x\) is defined component-wise as
\begin{equation}
    G_i(x)=
    \begin{cases}
        1, & \text{if } x \in [c_i-s,\; c_i+s],\\
        0, & \text{otherwise},
    \end{cases}
\end{equation}
for \(i=1,\dots,n_{\text{bin}}\), where \(n_{\text{bin}}\) denotes the encoding dimension. The bin centers \(c_i\) and window half-width \(s\) are given by
\begin{equation}
    c_i=\frac{2i-1}{2n_{\text{bin}}}, \quad i=1,\dots,n_{\text{bin}},
\end{equation}
\begin{equation}
    s=\frac{2\lfloor n_{\text{bin}}/2 \rfloor-1}{4n_{\text{bin}}}.
\end{equation}

In this work, we use \(n_{\text{bin}}=10\), corresponding to a 10-bit binary representation for each scalar input or reservoir value, with \(s=0.225\). This encoding strategy produces overlapping distributed binary patterns, ensuring that nearby continuous-valued inputs are mapped to binary vectors with small Hamming distances. As a result, basket encoding preserves local similarity structure in the binary embedding space while remaining fully compatible with the binary ON/OFF actuation of DMD micro-mirrors.

In addition to enabling efficient hardware-compatible encoding, the distributed and overlapping nature of basket encoding introduces redundancy in the binary representation, improving robustness against quantization artifacts and facilitating smoother state transitions in the reservoir dynamics.

\subsection*{Power-law Neuron Allocation Across Deep Reservoir Layers}
To investigate the effect of layer-wise neuron allocation in the proposed deep RC architecture, neurons are distributed across reservoir layers according to a normalized power-law decay. This strategy allocates a larger proportion of neurons to earlier layers and progressively fewer neurons to deeper layers, consistent with the hierarchical information flow of the architecture, where only the first reservoir layer directly receives the external input.

The layer-wise weighting profile is first defined using a power-law decay function:
\begin{equation}
\phi(l) = l^{-\gamma}, \quad l = 1,\dots,L,
\end{equation}
where \(L\) denotes the total number of reservoir layers and \(\gamma\) controls the decay rate of the neuron distribution. In all experiments, we use \(\gamma = 1.2\).
These weights are then normalized to preserve the total neuron budget:
\begin{equation}
w_l = \frac{\phi(l)}{\sum_{k=1}^{L}\phi(k)},
\end{equation}
such that \(\sum_{l=1}^{L} w_l = 1\).
Given a total neuron budget \(N\), the layer-wise neurons allocation is first computed as
\begin{equation}
\tilde{n}_l = N w_l.
\end{equation}
To ensure compatibility with the experimental encoding pipeline and facilitate efficient mapping onto the DMD, the number of neurons in each reservoir layer is constrained to be a multiple of 25. Accordingly, the neurons allocation is discretized as
\begin{equation}
n_l = 25 \left\lfloor \frac{\tilde{n}_l}{25} + \frac{1}{2} \right\rfloor, \quad l = 1,\dots,L.
\end{equation}
This rounding operation may introduce a small discrepancy with respect to the target total neuron budget. To preserve the exact neuron count, a global correction term is computed:
\begin{equation}
\Delta N = N - \sum_{l=1}^{L} n_l,
\end{equation}
and applied to the first layer as
\begin{equation}
n_1 \leftarrow n_1 + 25 \left\lfloor \frac{\Delta N}{25} + \frac{1}{2} \right\rfloor.
\end{equation}
The correction term is applied to the first reservoir layer, which typically contains the largest neuron allocation, thereby minimizing perturbations to the overall power-law distribution. This final adjustment ensures that the total neuron budget remains identical across all compared architectures while satisfying the hardware discretization constraints. Unless otherwise specified, all shallow and deep reservoir configurations are evaluated under the same fixed total neuron budget using this allocation strategy.

\subsection*{Benchmark tasks}
For each benchmark task described below, each input have specific preprocessing and are subsequently binarized by the basket encoding scheme described above. All target labels are represented using one-hot encoding, and classification is performed using ridge regression as described in the theoretical model section, with the regularization hyperparameter \(\lambda\) selected via Bayesian Optimization.

\paragraph*{Video Classification Task (KTH Dataset)}
The KTH human action recognition dataset is a standard benchmark for video-based action classification, containing six actions (\textit{walking}, \textit{running}, \textit{jogging}, \textit{boxing}, \textit{hand waving}, and \textit{hand clapping}) performed by 25 subjects under four recording scenarios. In this work, only the S1 outdoor scenario was considered, corresponding to 600 video sequences recorded at 25 frames per second.

For each video sequence, we perform frame extraction, convert them to grayscale, and normalized their intensity in the range [0,1]. Subsequently, we extract the Histogram of Oriented Gradients (HOG) descriptors~\cite{dalal2005histograms} for each individual frame to capture spatial and motion-related information. To reduce dimensionality, we apply Principal Component Analysis (PCA) and retain the first 1000 components per frame, corresponding to approximately 50 $\%$ of the cumulative explained variance.
 
To increase training diversity, each video sequence was divided into four equal temporal segments approximately capturing different temporal phases of the repeated action patterns within a recording. Each segment is then treated as an independent samples and are shuffled across action classes to generate temporally interleaved input streams and reduce overfitting to long recordings or subject-specific correlations. Finally, we evaluate the performance using 2-fold cross-validation with the two central segments used alternately for testing and the remaining three for training. Central segments were selected for testing because the beginning and end segments of a video sequence often contain entrance and exit frames with limited discriminative motion content.

\paragraph*{Image Classification Task (MNIST Dataset)}
The MNIST dataset contains $70,000$ grayscale images of handwritten digits (\(0\)–\(9\)) with a spatial resolution of $28 \times 28$ pixels. Since MNIST is a static image dataset, a preprocessing pipeline was used to convert each image into a sequential representation compatible with the temporally driven deep RC architecture. First,  images are normalized to the same intensity scale, then they are divided into four non-overlapping horizontal regions of 7 columns, from which Histogram of Oriented Gradients (HOG) features~\cite{dalal2005histograms} are extracted. This yields a sequence of four feature vectors per image, which preserves local spatial information while enabling sequential processing by the deep RC system. The extracted HOG descriptors were reduced using PCA to 25 principal components per input segment, preserving approximately 50 $\%$ of the cumulative variance. Finally, we concatenate the reservoir response of the four sequential input segments into a single feature vector to train the deep RC. For the performance evaluation, we did a $7$-fold cross validation where each fold was tested on $10,000$ images and trained on the remaining $60,000$ images to guarantee that each image was precisely tested one time.
 
\paragraph*{Audio Classification Task (TI-46 Dataset)}
The TI-46 spoken digit dataset consists of isolated utterances of the digits $0$-$9$ in $500$ audio recordings, with all ten digits being spoken ten times by five different speakers. Each audio recording is transformed into a two-dimensional time-frequency representation using Lyon's cochlear model, resulting in a biologically inspired cochleagram. This preprocessing keeps the fine grained temporal and spectral information.
 
Each cochleagram was zero-padded along the temporal dimension to a fixed size of $86$ frequency channels $\times$ $130$ time steps, to ensure uniform input dimensionality and avoid sequence-length bias during training. Then the cochleagrams were presented sequentially column-wise (\emph{i.e.} spectral activity at each time step). The dataset was further divided into groups of $50$ utterances, with each group having $5$ samples from each digit class, to ensure balanced evaluation folds. Within each group the utterances were shuffled randomly once before presentation, thus preventing the reservoir from processing long consecutive sequences of the same class while keeping balanced training and testing subsets. We evaluated performance using $10$-fold cross-validation, where each fold consists of $50$ utterances for the test set and the remaining $450$ for training.

\section*{Acknowledgments}
This work was supported financially by the Conseil of R\'{e}gion Grand-Est and by the European Office of Aerospace Research and Development (EAORD) and AirForce Office for Scientific Research (AFOSR) via Grant FA8655-22-1-7031. The authors acknowledge the support of the Chair in Photonics.

\section*{Author contributions statement}
D.R. designed the study. D.R. and N.M. managed the project and supervised M.W.I and M.A. M.W.I, N.M. and D.R. designed the experimental setup. All authors participated in the construction of the experimental setup. M.W.I performed the experiments, processed and analyzed the data. M.W.I, N.M and D.R. discussed and analyzed the results. M.W.I wrote the manuscript. N.M and D.R. presented suggestions for improving the quality of the work and revised the manuscript. M.W.I, N.M and D.R discussed and commented on the manuscript.

\section*{Conflict of interest}
The authors declare no competing interests.

\bibliography{Bibliography}
\end{document}